\documentclass{article}

\usepackage[utf8]{inputenc}
\usepackage[T1]{fontenc}
\usepackage{graphicx}
\usepackage{url}
\ifx\gabprop\undefined
\newcommand{\score}{\ensuremath{score}}
\newcommand{\State}{\ensuremath{state~s}}
\newcommand{\Action}{\ensuremath{action~a}}
\newcommand{\state}{\ensuremath{s}}
\newcommand{\Value}{\ensuremath{V}}
\newcommand{\action}{\ensuremath{a}}
\newcommand{\avg}{\ensuremath{avg~}}
\newcommand{\Reward}{\ensuremath{reward}}
\newcommand{\reward}{\ensuremath{r}}
\newcommand{\next}{\ensuremath{next}}
\newcommand{\numsims}{\ensuremath{num~sims}}
\newcommand{\uct}{\ensuremath{uct}}
\newcommand{\puct}{\ensuremath{puct}}
\newcommand{\Proba}{\ensuremath{Proba}}
\newcommand{\proba}{\ensuremath{p}}
\newcommand{\NN}{\ensuremath{NN}}
\else
\newcommand{\state}{\ensuremath{s}}
\fi

\begin{document}

\title{Polygames: Improved Zero Learning}


\author{Tristan Cazenave, Yen-Chi Chen, Guan-Wei Chen, \\Shi-Yu Chen, Xian-Dong Chiu, Julien Dehos,\\ Maria Else, Qucheng Gong, Hengyuan Hu,\\ Vasil Khalidov, Cheng-Ling Li, Hsin-I Lin,\\ Yu-Jin Lin, Xavier Martinet, Vegard Mella,\\ Jeremy Rapin, Baptiste Roziere, Gabriel Synnaeve,\\ Fabien Teytaud, Olivier Teytaud, Shi-Cheng Ye,\\ Yi-Jun Ye, Shi-Jim Yen, Sergey Zagoruyko}


\maketitle

\begin{abstract}
Since DeepMind's AlphaZero, Zero learning quickly became the state-of-the-art method for many board games. It can be improved using a fully convolutional structure (no fully connected layer).
Using such an architecture plus global pooling, we can create bots independent of the board size.
The training can be made more robust by keeping track of the best checkpoints during the training and by training against them.
Using these features, we release Polygames, our framework for Zero learning, with its library of games and its checkpoints.
We won against strong humans at the game of Hex in 19x19, which was often said to be untractable for zero learning;
and in Havannah. We also won several first places at the TAAI competitions.
\end{abstract}

\section{Introduction}
In spite of AlphaGo \cite{alphago}, some games still resist to computers and for many games the computational requirement is still huge. We present Polygames, our open-source zero learning framework available at \url{https://github.com/facebookresearch/polygames}.
It is based on Zero learning, combining Monte Carlo Tree Search and Deep Learning. 
It features a new architecture for accelerating the training and for making it size-invariant (Section \ref{struct}).
It allows neuroplasticity i.e. adding neutral layers, adding channels, increasing kernel size (Section \ref{dyn}) and warm start.
Polygames also features a new tournament mode for robustifying the training  (Section \ref{tournament}).
The framework provides a single-file API, that is generic enough for implementing many games, and comes with a library of games and many checkpoints.
Polygames made the first ever win against top level humans at the game of Hex 19x19 (Section \ref{vg}) and Havannah 8x8 (Section \ref{havsec}).

\subsection{Zero learning}
AlphaGo and AlphaZero \cite{alphago,alphazero} proposed a combination between Monte Carlo Tree Search \cite{coulom06} and Deep Learning.
The version in \cite{alphazero} is a simplified and elegant version, learnt end-to-end.
\subsubsection{Monte Carlo Tree Search}
Monte Carlo consists in approximating values (i.e. expected rewards) by averaging random simulations.
MCTS consists in biasing these random simulations: using the statistics from previous simulations: we increase the frequency of moves which look good. The most well known variant of MCTS is Upper Confidence Trees (UCT) \cite{uct}, which uses, for biasing the simulations, a formula inspired from the bandit literature: a move is chosen if it maximises a score as follows:
$$\score_{\uct}(\state, \action)=\avg\Reward(\next(\state,\action)) + k\sqrt{\frac{\log(\numsims(\state))}{\numsims(\next(\state,\action))}}.$$
An MCTS rollout consists in doing a simulation from the current board $s_0$ and playing using the above formula until a final state is reached.
When $M$ MCTS rollouts have been performed starting in the current state $\state_0$, we choose an action $\action$ maximizing e.g. $\numsims(\next(\state_0, \action))$. 

\subsubsection{Neural policies}
Let us assume that we have a function $tensor$ which maps a state to a tensor $tensor(state)$.
A neural network (NN) can then be applied. We consider a NN with two outputs:
\begin{eqnarray*}
\pi_{NN}(state)& &\mbox{ is a tensor,}\\
V_{NN}(state)& &\mbox{ is a real number.}
\end{eqnarray*}
The NN typically uses convolutional layers, and then two heads for those two outputs; it originally contains fully connected layers in both heads.
If we assume that each possible action has an index in the output tensor, then
$\pi_{\NN}(\state)$ can be converted into probabilities of actions by (i) multiplying by some temperature $T$ and applying the $exp$ function to each entry in it (ii) setting to zero illegal actions in $\State$ (iii) dividing by the sum. Let us note $\Proba_{\NN}(\state,\action)$ the probability of action $\action$ in state $\state$ for the neural network $\NN$. Then, an action can be sampled.

\subsubsection{Zero model of play: how the MCTS is modified by adding the NN}
The zero-model is then as follows.
First, the UCT formula is adapted as PUCT \cite{alphazero}, as follows:
$$\score_{\puct}(\State, \Action)=\avg\Reward(\next(s,a)) + \Proba_{\NN}(\state,\action)\sqrt{\frac{\numsims(\state)}{\numsims(\next(\state,\action))}}.$$
Please note that the $\log$ has been removed. The parameter $k$ has been replaced by the probabilities provided by $\NN$.

Second, when a simulation reached a state in which no statistics are available (because no simulation has been performed here), instead of returning the reward of a random rollout until the end of the game, we return the reward estimated by $\Value_{\NN}$. The NN has therefore the double impact of (i) biasing the tree search (ii) truncating Monte Carlo simulations.

\subsubsection{Zero training: how the NN is modified by the MCTS}\label{zerotrain}
Let us generate games with a zero-model as above, using e.g. $M=600$ simulations per move.
Each time a game is over, we have a family of 3-tuples $(\state,\proba,\reward)$, one per visited state in this game.  $\reward$ is the final reward of the game, and $\proba_{location_\action}$ is the tensor of the proportion of the zero-simulations using action $\action$ in state $\state$, and $location_\action$ is the index of $\action$ in the output tensor.
These 3-tuples are then used for training the network so that $\pi_{\NN}$ imitates $\proba$ (cross-entropy) and $\Value_{\NN}$ approximates the reward ($l^2$ loss). We also use a weight decay as a regularization.

\subsubsection{Overall architecture}
There is a server and many clients:
\begin{itemize}
	\item The server receives 3-tuples $(\state,\proba,\reward)$ from the clients. It stores them in a replay buffer, in a cyclic manner. It trains the NN as in Section \ref{zerotrain}, also cycling over the replay buffer.
	\item The clients send data (3-tuples) to the server.
\end{itemize}
There are other subtleties, such as adding a Dirichlet noise for more diversity. The number of clients should be tuned so that the cycles performed by the trainer are just a bit faster than the speed at which data are provided; Section \ref{other} provides a robust solution for ensuring this, in particular for low computational power.

\subsection{Other open-source frameworks}

Many teams have repeated and sometimes improved the Alpha Zero approach for different games.

Elf/OpenGo \cite{tian2019elfopengo} is an open-source implementation of AlphaGo Zero for the game of Go. After two weeks of training on 2,000 GPUs it reached superhuman level and beat professional Go players.

Leela Zero \cite{pascutto2017leela} is an open-source program that uses a community of contributors who donate GPU time to replicate the Alpha Zero approach. It has been applied with success to Go and Chess.

Crazy Zero by Rémi Coulom is a zero learning framework that has been applied to the game of Go as well as Chess, Shogi, Gomoku, Renju, Othello and Ataxx. With limited hardware it was able to reach superhuman level at Go using large batches in self-play and improvements of the targets to learn such as learning territory in Go. Learning territory in Go increases considerably the speed of learning.

KataGo \cite{greatfast} is an open-source implementation of AlphaGo Zero that improves learning in many ways. It converges to superhuman level much faster than alternative approaches such as Elf/OpenGo or Leela Zero. It makes use of different optimizations such as using a low number of playouts for most of the moves in a game so as to have more data about the value in a shorter time. It also uses additional training targets so as to regularize the networks.

Galvanise Zero \cite{gzero} is an open-source program that is linked to General Game Playing (GGP) \cite{Pitrat68}. It uses rules of different games represented in the Game Description Language (GDL) \cite{love2008general}, which makes it a truly general zero learning program able to be applied as is to many different games. The current games supported by Galvanise Zero are Chess, Connect6, Hex11, Hex13, Hex19, Reversi8, Reversi10, Amazons, Breakthrough, International Draughts.

\section{Innovations}\label{inno}
\subsection{Structured zero learning}\label{struct}
\subsubsection{Fully convolutional models}\label{fc}
Many zero-learning methods are based on traditional convolutions, followed by fully connected layers. However, policy learning in board games is often closer to image segmentation than to classical image classification as actions are naturally mapped on boards. The input has various channels, and two dimensions matching the board size. Similarly, the output of the network has various channels, corresponding to various possible moves, and two dimensions matching the board size as well. Therefore, we can apply fully convolutional models - not a single fully connected layer is necessary in zero learning, and such a layer perturbates the learning. 

\subsubsection{Scale invariant models}
As usually in zero learning, our neural network has two heads: one for the policy and one for the value. The one for the policy is fully convolutional (Section \ref{fc}), and therefore it works independently of the input size, i.e. independently of the board size.
The value part, however, does not have this property if it is fully connected. We therefore use global pooling as in \cite{greatfast}.
Global pooling replaces each channel $c$, of shape possibly $boardsize\times boardsize\times 1$, by several channels, such as the maximum and the average of $c$, over the $boardsize \times boardsize$ entries. We therefore get a boardsize-independent representation.
Our Hex19 model was trained in $13 \times 13$ and was immediately strong in $19 \times 19$ -- though we needed a bit of fine tuning for the success story presented in Section \ref{vg}.

\subsection{Neuroplasticity}\label{dyn}
Several modifications are almost neutral when initialized close to zero: 
\begin{itemize}
	\item addition of residual blocks (i.e. switching from 12 blocks of 3 convolutional layers to 13 or 14 blocks of 3 convolutional layers);
	\item addition of new channels;
	\item extension of the kernel size (from $3 \times 3$ to $5 \times 5$ or $5 \times 5$ to $7 \times 7$, etc).
\end{itemize}
Polygames provides a script ``convert'' that makes such a growth of the neural network easy. Training can be resumed after such extensions of the neural architectures; we can train, then grow (while preserving the quality of the model as it remains almost equal to the previous model as new weights are close to 0), then resume the training with more degrees of freedom.

\subsection{Tournament mode}\label{tournament}
In order to fight catastrophic forgetting or the red queen effect (oscillations of performance), we add a tournament mode: at each instant, ten models are kept in the ranking; as their ELO rating is computed from their results against the current model; each time a new model is saved, we remove the worst model from the pool. Each client plays games between the current model (termed ``dev'') and a model with a given ELO selected with probability proportional to $\exp(-\frac{ELO_{dev}-ELO}{400})$.

\subsection{Other features}\label{other}

 We provide  checkpoints  \footnote{\url{http://dl.fbaipublicfiles.com/polygames/checkpoints/list.txt}}
 for many games: Einstein W\"urfelt Nicht, Breakthrough, Havannah8, Havannah10, MiniShogi, Othello8, Othello10...

 Heuristically, we consider that an example, in the replay-buffer, should never be seen more than 8 times. When the clients are not fast enough for filling the replay buffer, for example because of preemption of clients or slow game,  we artificially add delays in the learning.	

 Adding a new game can be made by writing a new class that inherits from State and overrides a few methods (see the implementation of Connect Four \footnote{\url{https://github.com/facebookresearch/polygames/blob/master/games/connectfour.h}} as an example).

 The code can handle stochastic games; our bot ``randototoro'' performs quite well on LittleGolem at the game of Einstein W\"urfelt Nicht.

 One can also add one-player games. Examples include Mastermind, Minesweeper, and various combinatorial optimization problems.
 
 MCTS can be adapted easily for one-player games, and in a more tricky manner to partially observable games in which the visible information is the same for both player (trivially for Chinese Dark Chess, but also for Mastermind and Minesweeper). We adapt the method used in \cite{pomcts} to the Zero setting.

\section{Success stories}\label{ss}
\subsection{Beating humans at Hex19}\label{vg}
\begin{figure}
\center
	\includegraphics[width=.45\textwidth]{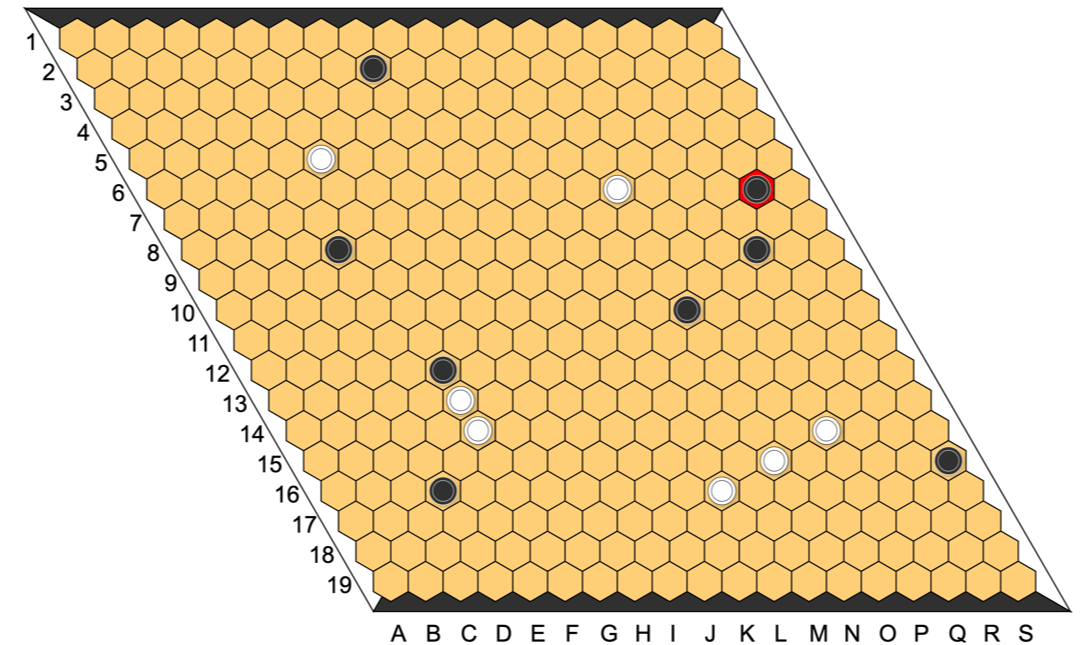}
	\includegraphics[width=.45\textwidth]{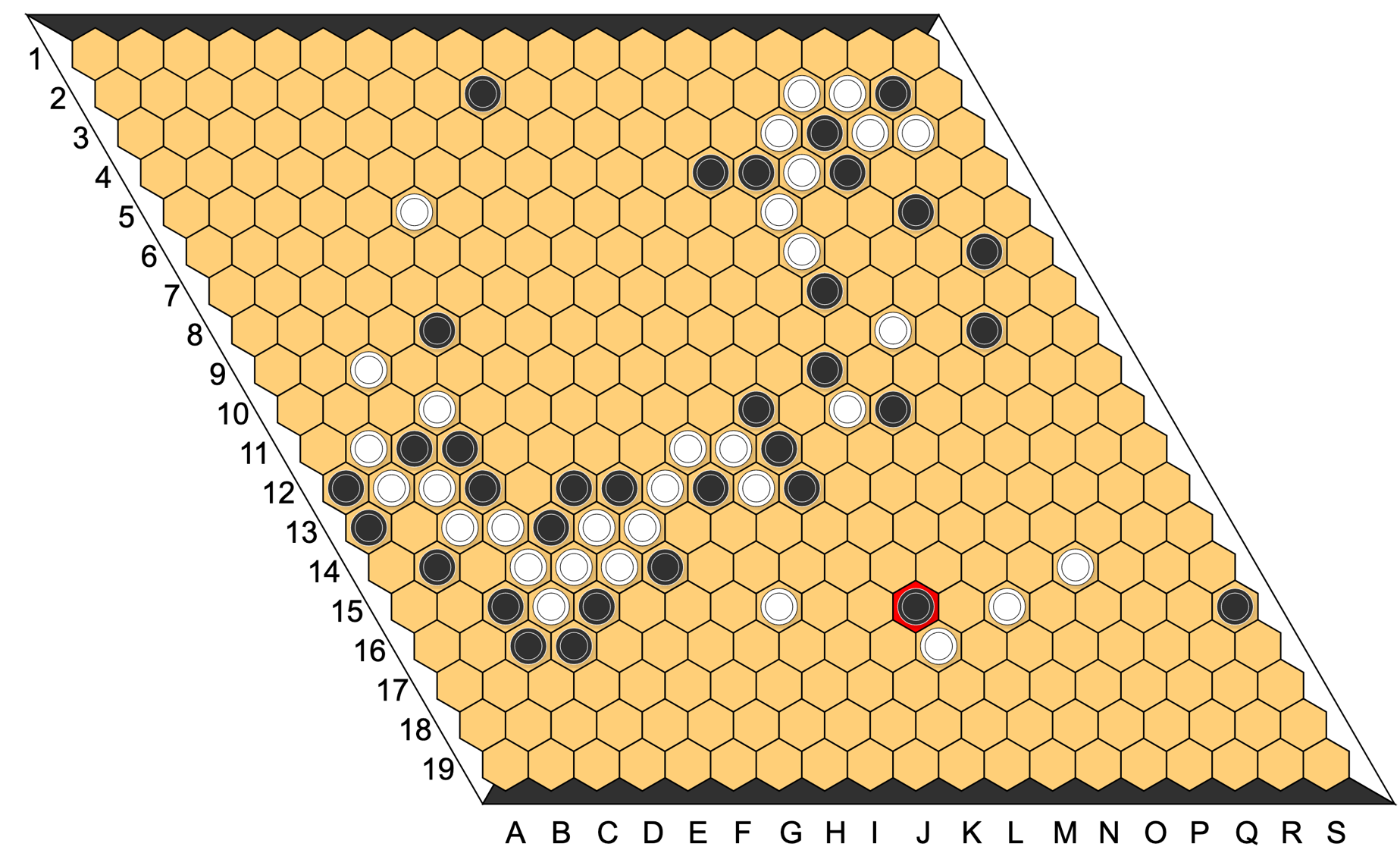}\\
	\includegraphics[width=.45\textwidth]{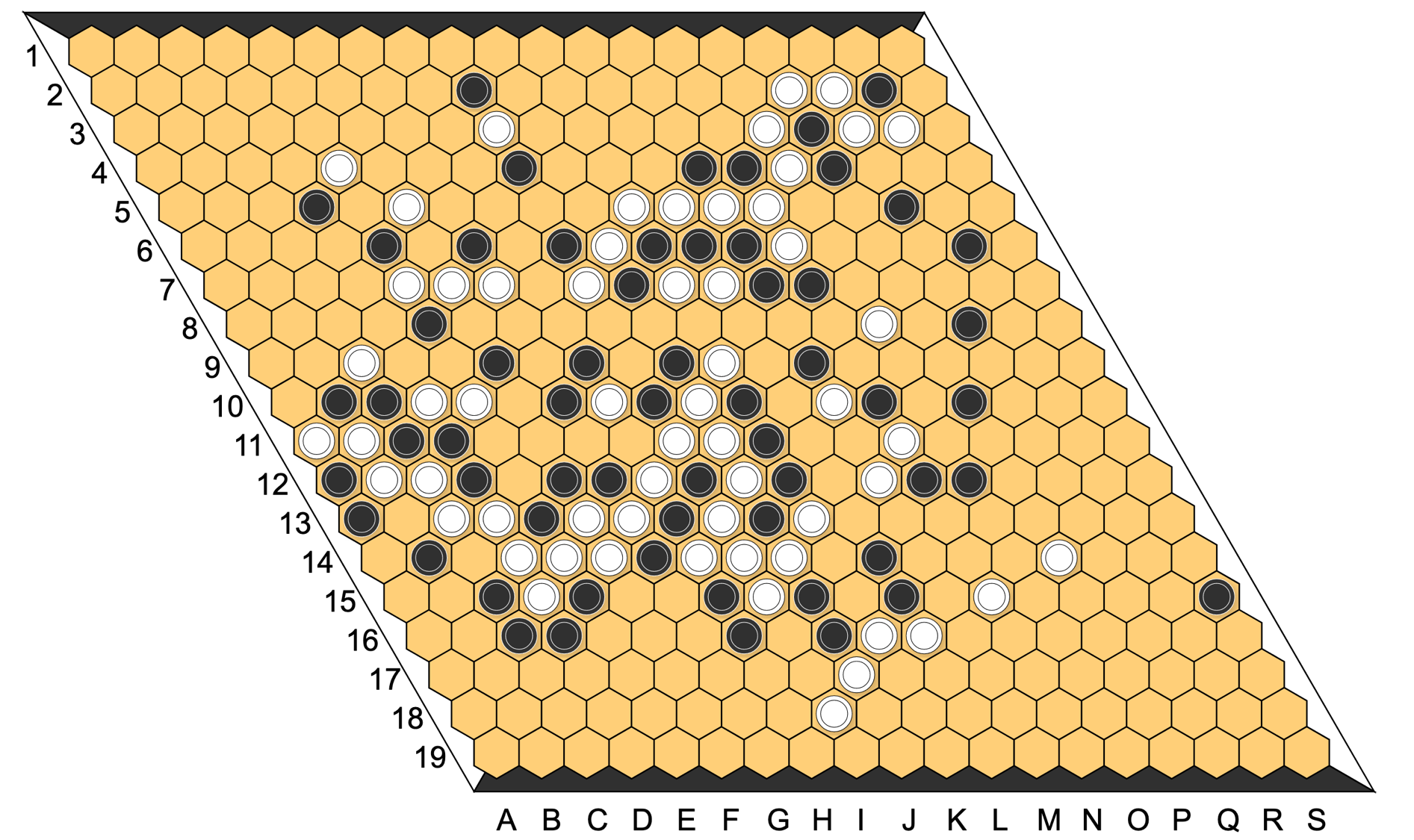}
	\includegraphics[width=.45\textwidth]{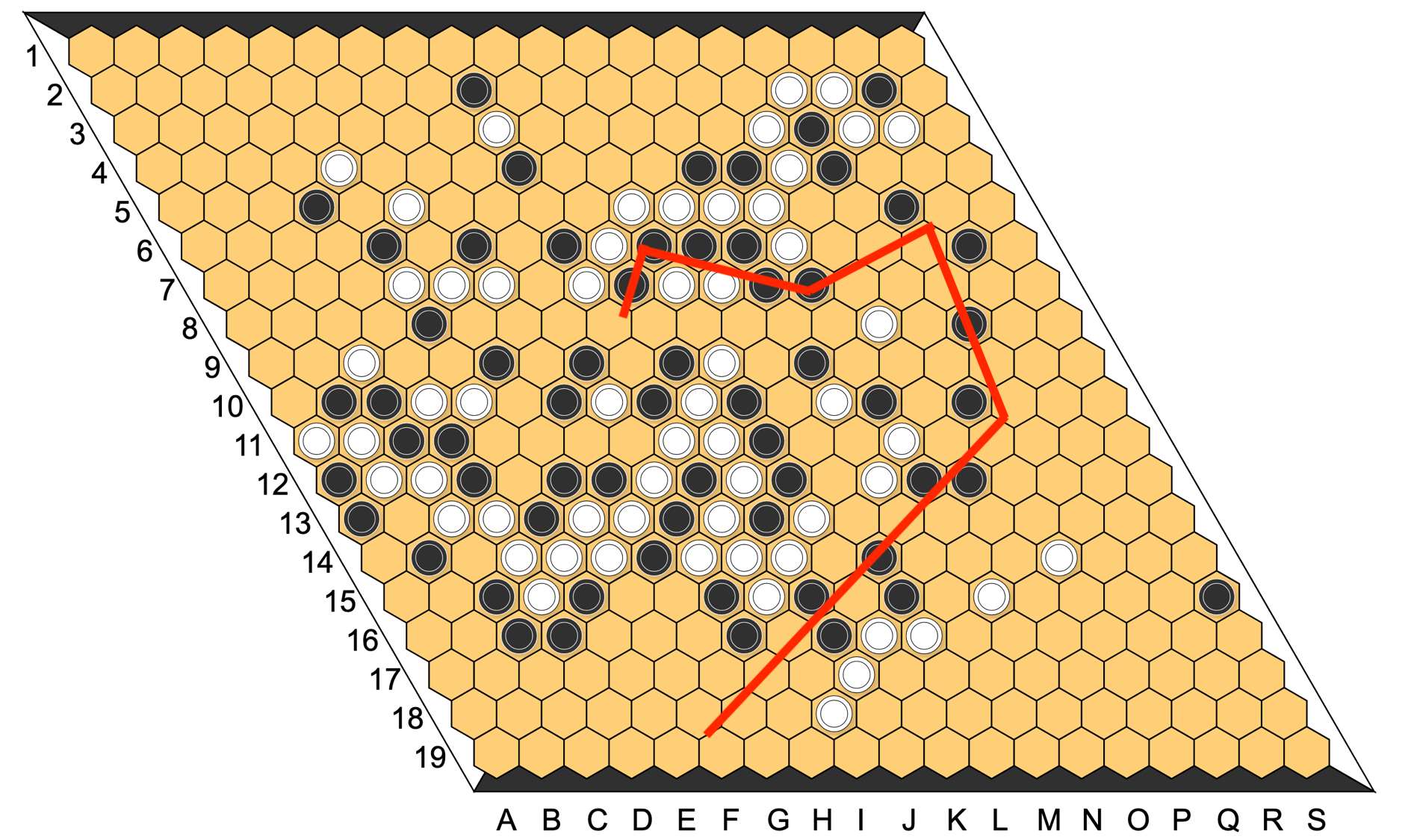}\\
	\caption{\label{hex}Game of Hex 19x19 with Pie rule played by Polygames against Arek Kulczycki, Winner of the last LG tournament, ranked first on the LittleGolem server. First: opening. Second: at that stage, the human (White) seems to win - two solid groups are connected to East and West respectively, and look close to connect each other. However (third), the situation in the center is quite complicated and later it turns out that Black can win by one of two possible paths (last: White can not block both H7 and an attack on the left side). Source: LittleGolem.}
\end{figure}

According to \cite{bonnet}, “Since its independent inventions in 1942 and 1948 by the poet and mathematician Piet Hein and the economist and mathematician John Nash, the game of hex has acquired a special spot in the heart of abstract game aficionados. Its purity and depth has lead Jack van Rijswijck to conclude his PhD thesis with the following hyperbole \cite{jack}: << Hex has a Platonic existence, independent of human thought. If ever we find an extraterrestrial civilization at all, they will know hex, without any doubt.>> ”
The rules are simple. Black and white fill an empty cell, in turn (Fig. \ref{hex}). Black wins if it connects North and South, White wins if it connects West and East.
The game is made more fair by a pie rule: at the second move, the second player can decide to swap colors.
The game is hard because, as a connection game, its reward is based on a global criterion (no local criterion to sum). Fig. \ref{hex} shows the win against humans.

\subsection{TAAI competition}
At TAAI 2019, Polygames was ranked first in Othello10, Breakthrough and Connect6 \footnote{\url{https://www.tcga.tw/taai2019/en/}}.
For more statistical significance, it was also successfully tested against WZebra and Ltbel (Othello8), the winner of TAAI2018 at Connect6 and won all games.
\def\maybelater{\subsection{Einstein W\"urfelt Nicht: a stochastic game}
We use two levels of nodes, one for decision nodes and one for random nodes.
Our bot reaches strong level on LittleGolem. }
\subsection{Havannah}\label{havsec}
\begin{figure}
	\includegraphics[width=.3\textwidth]{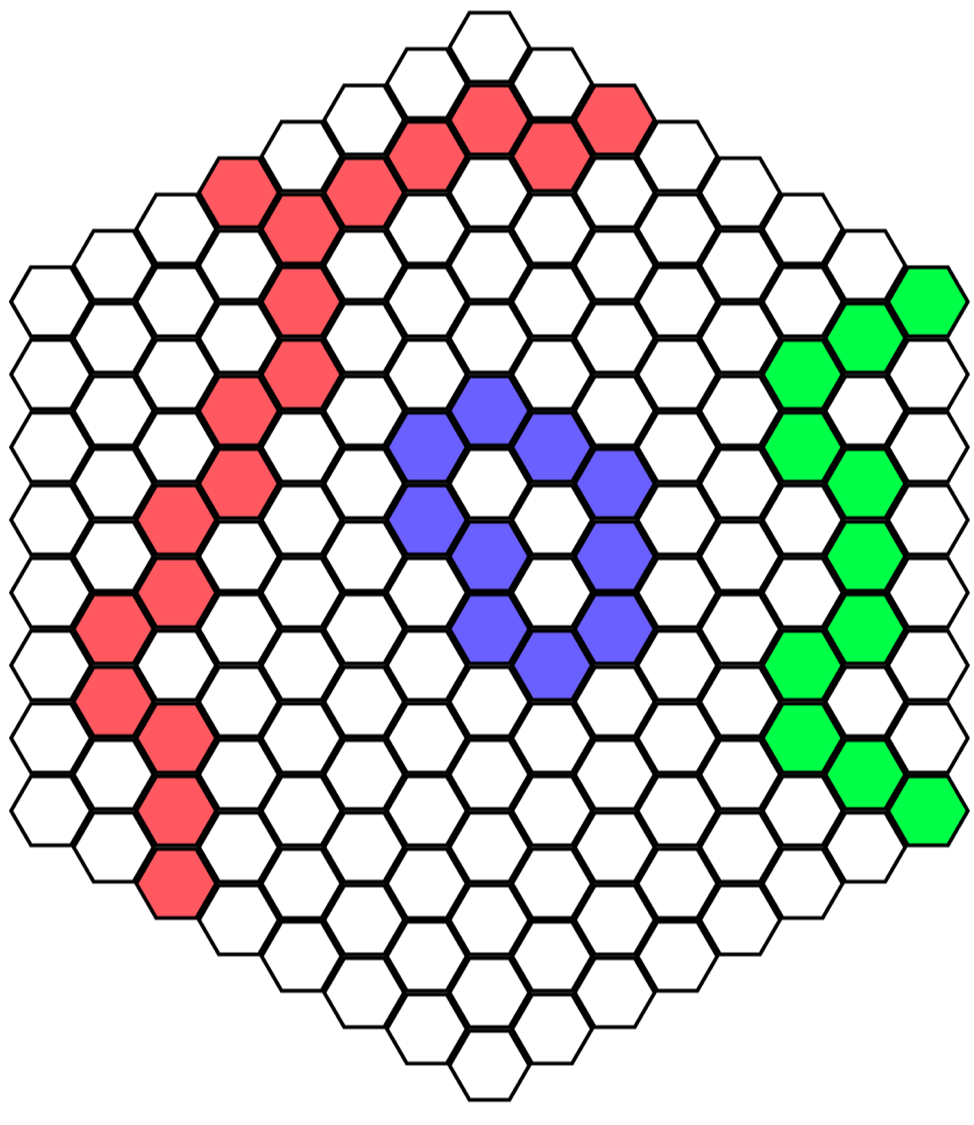}
	\includegraphics[width=.3\textwidth]{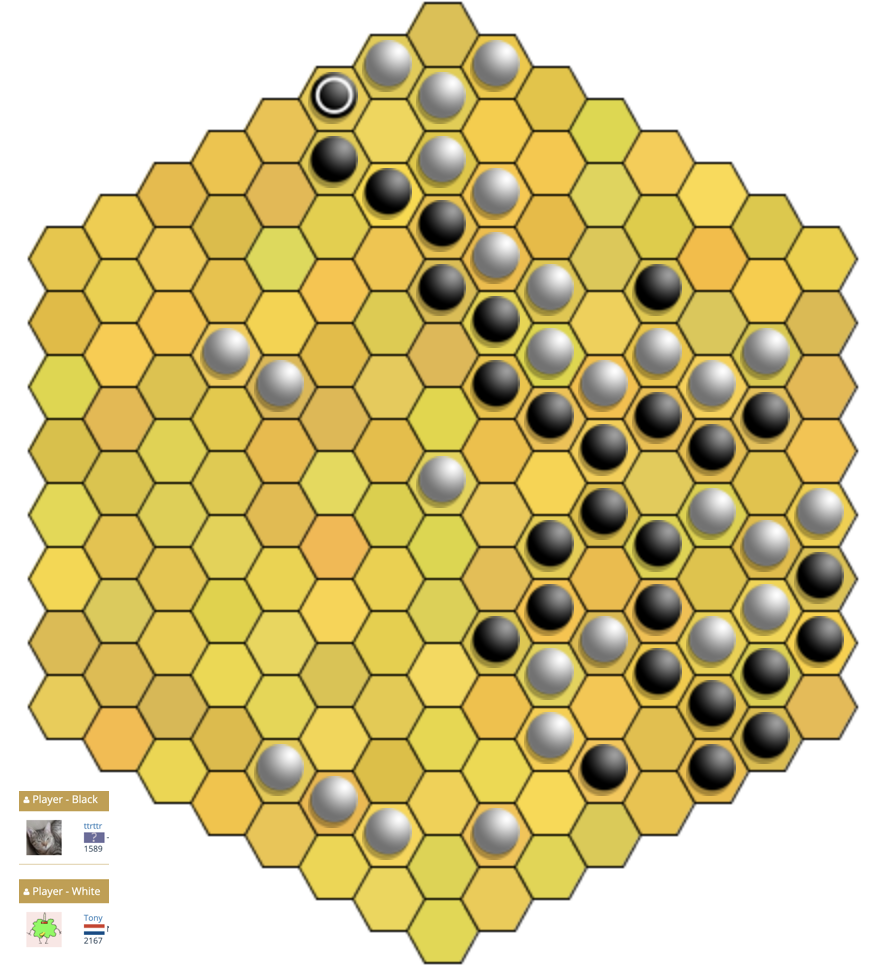}
	\includegraphics[width=.3\textwidth]{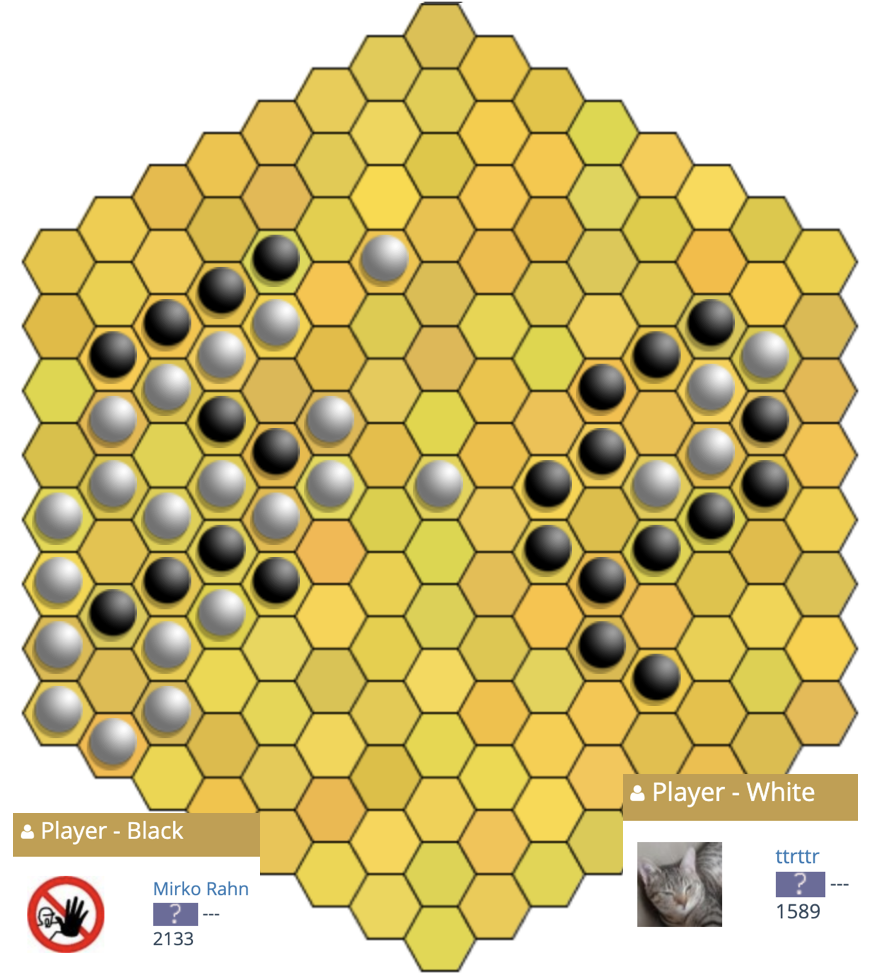}
	\caption{\label{hav}Left: The game of Havannah and the three different ways for winning. Middle: a win by Polygames against Tony, Elo rank 2167. Right: a win against Mirko Rahn, Elo rank 2133, winner of many Havannah tournaments. Sources: wikipedia (left) and LittleGolem (middle and right).}
\end{figure}

Havannah was invented specifically for being hard for computers (Fig. \ref{hav}).
It follows the game play of Hex, also with hexagonal cells but now on an hexagonal board, and winning conditions are more diverse:
\begin{itemize}
	\item Connecting two of the six corners wins (15 possibilities);
	\item Connecting three of the six sides wins (20 possibilities);
	\item Realizing a loop (even if it does not contain empty cells) wins.
\end{itemize}
According to \cite{lorentz}, ``The state of Havannah programming is still in its early stages. Though the top programs do play at a reasonable level, about the level of somebody who has played the game for 6 months or a year, they still play with a very unnatural style, and often win their games by virtue of tactical shots missed by the human opponent. Our feeling is that Havannah programs cannot be expected to play at an elite level until they learn to play a more natural, human-like game.''

Draws are theoretically possible but very rare. The game is also played with a pie rule. Some decent players have been defeated by computers, but never the best humans until Polygames. On Littlegolem, we have won 3 games out of 4 against \textit{Mirko Rahn} (Elo rank 2133), and 2 games out of 2 against \textit{tony} (Elo rank 2167), who belong to the top four players on this website (see Figure~\ref{hav} middle and right for examples of games).

\section{Conclusions}

We propose a state-of-the-art framework, called Polygames, that can play to various games. It is based on zero learning, with innovations detailed in Section \ref{inno}, and had success stories detailed in Section \ref{ss}.
Polygames contains new architectures, allows architecture search thanks to neutral components addition and is stabilized by a tournament mode and an overfitting fighting method.
It was widely tested in the TAAI competition and on LittleGolem (\url{www.littlegolem.net}). The source code is publicly available under an open-source license.

\bibliographystyle{abbrv}  
\bibliography{z,rl}

\begin{thebibliography}{10}

\bibitem{bonnet}
{\'{E}}.~Bonnet, F.~Jamain, and A.~Saffidine.
\newblock On the complexity of connection games.
\newblock {\em Theor. Comput. Sci.}, 644:2--28, 2016.

\bibitem{pomcts}
O.~Buffet, C.-S. Lee, W.~Lin, and O.~Teytaud.
\newblock {Optimistic Heuristics for MineSweeper}.
\newblock In {\em {International Computer Symposium}}, Hualien, Taiwan, 2012.

\bibitem{coulom06}
R.~Coulom.
\newblock Efficient selectivity and backup operators in {Monte-Carlo} tree
  search.
\newblock In {\em Proceedings of the 5th International Conference on Computers
  and Games}, CG'06, pages 72--83, Berlin, Heidelberg, 2007. Springer-Verlag.

\bibitem{gzero}
R.~Emslie.
\newblock Galvanise zero.
\newblock \url{https://github.com/richemslie/galvanise_zero}, 2019.

\bibitem{uct}
L.~Kocsis and C.~Szepesv{\'a}ri.
\newblock Bandit based {Monte-Carlo} planning.
\newblock In {\em Machine Learning: ECML 2006}, pages 282--293. Springer, 2006.

\bibitem{lorentz}
R.~Lorentz.
\newblock Early playout termination in mcts.
\newblock In A.~Plaat, J.~van~den Herik, and W.~Kosters, editors, {\em Advances
  in Computer Games}, pages 12--19, Cham, 2015. Springer International
  Publishing.

\bibitem{love2008general}
N.~Love, T.~Hinrichs, and M.~Genesereth.
\newblock General game playing: Game description language specification.
\newblock {\em Stanford Logic Group Computer Science Department Stanford
  University}, 2006.

\bibitem{pascutto2017leela}
G.-C. Pascutto.
\newblock Leela zero.
\newblock \url{https://github.com/leela-zero/leela-zero}, 2017.

\bibitem{Pitrat68}
J.~Pitrat.
\newblock Realization of a general game-playing program.
\newblock In {\em Information Processing, Proceedings of {IFIP} Congress 1968,
  Edinburgh, UK, 5-10 August 1968, Volume 2 - Hardware, Applications}, pages
  1570--1574, 1968.

\bibitem{alphago}
D.~Silver, A.~Huang, C.~J. Maddison, A.~Guez, L.~Sifre, G.~van~den Driessche,
  J.~Schrittwieser, I.~Antonoglou, V.~Panneershelvam, M.~Lanctot, S.~Dieleman,
  D.~Grewe, J.~Nham, N.~Kalchbrenner, I.~Sutskever, T.~Lillicrap, M.~Leach,
  K.~Kavukcuoglu, T.~Graepel, and D.~Hassabis.
\newblock Mastering the game of {Go} with deep neural networks and tree search.
\newblock {\em Nature}, 529(7587):484--489, 2016.

\bibitem{alphazero}
D.~Silver, T.~Hubert, J.~Schrittwieser, I.~Antonoglou, M.~Lai, A.~Guez,
  M.~Lanctot, L.~Sifre, D.~Kumaran, T.~Graepel, T.~P. Lillicrap, K.~Simonyan,
  and D.~Hassabis.
\newblock Mastering chess and shogi by self-play with a general reinforcement
  learning algorithm.
\newblock {\em CoRR}, abs/1712.01815, 2017.

\bibitem{tian2019elfopengo}
Y.~Tian, {Jerry Ma*}, {Qucheng Gong*}, {Shubho Sengupta*}, Z.~Chen,
  J.~Pinkerton, and C.~L. Zitnick.
\newblock Elf opengo: An analysis and open reimplementation of alphazero.
\newblock {\em CoRR}, abs/1902.04522, 2019.

\bibitem{jack}
J.~van Rijswijck.
\newblock Set colouring games.
\newblock {\em Ph.D. thesis, University of Alberta}, 2006.

\bibitem{greatfast}
D.~J. Wu.
\newblock Accelerating self-play learning in go.
\newblock {\em CoRR}, abs/1902.10565, 2019.

\end{thebibliography}
\end{document}